# ReNN: Rule-embedded Neural Networks

Hu Wang
Data Analysis and Algorithm Department
LOHAS Technology (Beijing) Corporation Limited
Beijing, China
Wlys111@163.com

*Abstract*—The artificial neural network shows powerful ability of inference, but it is still criticized for lack of interpretability and prerequisite needs of big dataset. This paper proposes the *Rule-embedded Neural Network* (ReNN) to overcome the shortages. ReNN first makes local-based inferences to detect local patterns, and then uses rules based on domain knowledge about the local patterns to generate rule-modulated map. After that, ReNN makes global-based inferences that synthesizes the local patterns and the rule-modulated map. To solve the optimization problem caused by rules, we use a two-stage optimization strategy to train the ReNN model. By introducing rules into ReNN, we can strengthen traditional neural networks with long-term dependencies which are difficult to learn with limited empirical dataset, thus improving inference accuracy. The complexity of neural networks can be reduced since long-term dependencies are not modeled with neural connections, and thus the amount of data needed to optimize the neural networks can be reduced. Besides, inferences from ReNN can be analyzed with both local patterns and rules, and thus have better interpretability. In this paper, ReNN has been validated with a time-series detection problem.

*Keywords—neural network; feature mapping; rule embedding; knowledge representation; interpretability*

## Introduction

With decades of development, the artificial neural network (ANN) now shows powerful ability of inference, approaching or even surpassing human level in several specific scenarios such as image recognition [1, 2] and chess games [3]. One reason for such success is that the ANN could manage a great number of parameters with deep networks to learn highly complex functions. However, the interpretability of ANN is often criticized since ANN inferences are difficult to be explained as concise interaction among the parameters and the network.

Several researches have attempted to open the "black box" of ANN and to interpret its inferences. How to get insight into the features generated automatically in the hidden layer of ANN is a key issue for opening this "box". Schwartz-Ziv and Tishby proposed to visualize ANN with the Information Plane, and revealed the function of efficient representation of the hidden layers [4]. Zeiler *et al.* investigated the activations of neural units in feature layer of ANN, and highlighted the regions of the input data which are responsible for the activations [5, 6]. This approach is useful to dissect ANN models and to suggest ways to improve them [5]. Szegedy *et al.* analyzed ANN models by synthesizing samples that lead to high activations of the neural units [7, 8], and found that ANN models are easily hacked by adding certain structured noise in image space, revealing that supervised learning of ANN models might ignore some common sense of the target, e.g. mammals have four legs generally.

Recently, neural units in ANN feature layer have been interpreted with human-interpretable semantic concepts with the help of a densely labeled dataset [9, 10], or modeled as an explanatory graph which reveals the knowledge hierarchy hidden inside a pre-trained model of ANN [11]. Although these techniques have improved the interpretability of ANN, they cannot interpret an inference as concisely as human, e.g. telling why a sample is a dog rather than a cat. The main reason is that ANN is hard to learn knowledge of common sense from a single limited empirical dataset.

Compared with ANN, human make inferences as well as interpret inferences concisely based on two foundations, i.e. experiences and knowledge. With knowledge that learned and summarized from the development of science and technologies, human can make better inferences with a smaller empirical dataset. Therefore, it is vital to "teach" ANN models knowledge to improve ANN in several aspects: reducing the model complexity and cost to learn a new problem, adding interpretability with knowledge and increasing accuracy.

Knowledge representation have been used to describe the richness of the world in computer systems, which can be understood by artificial intelligence and used in reasoning and inferring subsequently [12, 13]. Rule-based representation is one of the most common formalisms of knowledge representation. For example, expert systems based on a set of rules have been widely used in computer-aided medical diagnosis [14, 15], data processing and analyzing [16, 17], fault diagnosis [18], etc.

The subsequent question is how knowledge representation could be combined with ANN. The knowledge-based ANN is an answer which designed the topological structure of ANN according to the knowledge of input variables and represented the knowledge as dependency structures of rules [19-21]. However, the answer is not appropriate for the deep ANN whose input variables are often high-dimensional raw data while the designers of ANN often have little idea about the dependency structures of high-dimensional raw data.

This paper proposes the Rule-embedded Neural Network (ReNN) which makes use of knowledge to improve the performance of ANN. Rules are used to represent domain knowledge and common sense in a computable manner. ReNN disassembles the "black box" of ANN into two parts: local-based inference and global-based inference. The local-based inference handles local patterns intuitively which can be easily learned from empirical datasets, while the global-based inference introduces rules about local patterns which have been accumulated by human for long time. Accordingly, knowledge from human teachers and experiences from empirical datasets are combined, and then contribute to the global-based inference together to improve the performance. Besides, we can differentiate the contributions of local patterns and rules to the final inferences, and thus improve the interpretability of the neural networks. As an example, we apply ReNN to a time-series detection problem in the experiments.

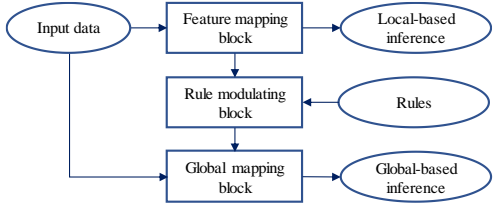

Figure 1. Computational graph of ReNN

## METHOD

### A. Basic computation graph

The Rule-embedded Neural Network (ReNN) is defined as:

$$\text{ReNN} = \{f, r, g\}, \tag{1}$$

where $f$, $r$ and $g$ represent the feature-mapping block, rule-modulating block, and global-mapping block, respectively. The computational graph for ReNN is shown in Fig. 1.

Given a supervised-learning problem with input data $X$ and targets $Y$. ReNN first extracts feature map from the input data with the feature-mapping block:

$$F = f(X), \tag{2}$$

where the output feature map $F$ is designed to reflect features of the targets. The features could be subcomponents of the targets. For example, a task is to detect all face occurrences with a two-dimensional image, and the feature map could consist of neural activations of face organs, i.e. eyes, ears, nose, and mouth and so on. The features could also be instances of the targets. For example, a task is to detect all R-peaks from electrocardiograph (ECG) of 60 seconds, and the feature map could consist of neural activations of multiple R-peaks.

Then, ReNN applies rules to the feature map with the rule-modulating block, and generates a rule-modulated map.

$$R = r(F) \tag{3}$$

The rules are designed according to domain knowledge, activating nodes in the rule-modulated map if and only if rule-based evidences could be found with the activations of the feature map. In the scenario of face detection, the relative positions of the face subcomponents can be utilized to analyze candidate positions and scales of face occurrences. In the scenario of R-peak detection from ECG, periodicity and variability of heart rate can be used to estimate possibility of R-peak-occurrences at candidate positions.

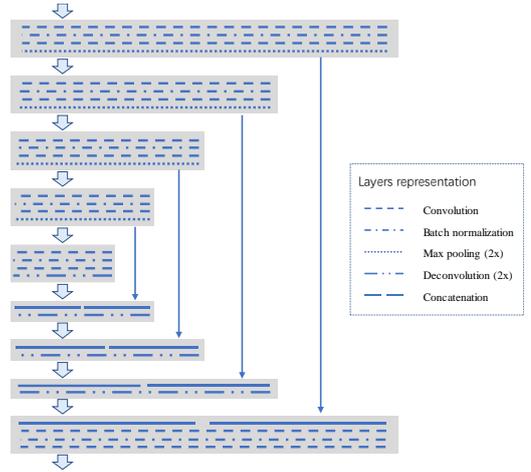

Figure 2. Block design for the feature-mapping and the global-mapping with FCN architecture for time-series data. It consists of 12 convolution layers, 6 batch normalization layers, 4 max pooling (2×) layers, 4 deconvolution layers (2×), and 4 concatenation layers, which are ordered as the hollow arrows show. The filter size of last convolution layer is 1×1, while others are 1×3. The solid arrows show that shallower features from the output of batch normalization layers are fused to deeper layers with the concatenation layers.

With the rule-modulated map and the input data, ReNN generates estimation of targets as follows:

$$\hat{Y} = g(X, R) \tag{4}$$

The feature map can $F$ also be used as an input of $g$ to accelerate the model training procedure.

Rule-modulating plays an important role for global mapping in ReNN. However, rules may make the ReNN model non-differentiable, leading to the failure of model training with gradient-descend-based optimization methods. A two-stage optimization strategy is adopted to overcome this limitation. The feature-mapping block is firstly optimized with a training dataset of $\{X, F\}$, and then the global-mapping block is optimized with dataset $\{X, Y\}$ while fixing the feature-mapping block. Same as traditional ANN, gradient-descend-based optimization methods are used at each stage of the strategy [22].

### B. Neural Block design

To further design the ReNN blocks, we should consider the characteristics of the specific task, such as the data dimension and target scales. The task of R-peak detection from one-dimensional ECG will be considered as an example in the following sections. ECG consists of time-series data of the electrical activity of the heart, which are recorded by noninvasive electrodes placed on the skin. The R-peak of ECG is a key time point in the procedure of rapid depolarization of the right and left ventricles. R-peaks are very useful when we need to estimate the exact time of each heartbeat, e.g. calculating heart rate variability [23] and pulse wave transit time [24, 25]. The sampling rate of ECG is 125 Hz in this paper, and the scale of the R-peak morphology is about 80 time-points (0.64 seconds).

We use the fully-convolutional network (FCN) [26] to design the feature-mapping block and the global-mapping block.

The FCN architecture for time series data is shown in Fig. 2. The convolution layer is used to capture local patterns, while the max-pooling layer is used to summarize local patterns and enlarge the receptive field of a neural note in deeper layers. The batch normalization layer is used to overcome the vanishing gradient problem and accelerate the training of deep neural networks. The deconvolution layer is used to upsample the pooled data, so that pointwise classification can be achieved in one feedforward computation [26]. The concatenation layer is used to reserve fine features from shallower into deeper layers where features are coarser due to the pooling operations.

As we know, a neural node after two layers of 1×3 convolution has a receptive filed of five, and the max pooling and deconvolution layers can double the receptive field. In sum, each output node in the FCN architecture has a receptive field of about 80 time-points (0.64 seconds). This receptive field covers all local information of the R-peak morphology, i.e. P-Q-R-S-T patterns of the electrical activity of a heartbeat. The receptive field can be enlarged with more max-pooling layers.

The number of output channels of the convolution layers and deconvolution layers are important hyper-parameters to control the complexity of the feature layers. We set output channels of all layers to be the same in the experiments for simplicity.

### C. Rule-modulating block

The rule-modulating block is designed according to our knowledge about R-peaks in ECG. To distinguish R-peaks from noises, human beings first model a piece of ECG signals with the knowledge of heart beats and ask questions. For example, is the owner of the ECG with normal sinus rhythm or with arrhythmias? What is the heart rate? If he or she is with normal sinus rhythm, the heart rate would be used to pick R-peaks out from the noises according to the time distance from neighboring R-peaks. The knowledge can be represented as rules.

The first step is to analyze some knowledge-based concepts from the feature map. The concepts here are heart rate (HR) and its standard deviation (SDNN). With the help of local-based inference, most of the R-peaks are prominent and easy to detect in the feature map with the Softmax function. We refer the set of time intervals of the R-peaks as Z, where a constraint from 0.3 to 1.5 seconds is applied to the intervals according to domain knowledge about heart rhythm of common users. With the assumption of Gaussian distribution for Z, we can eliminate abnormal intervals caused by noises. After that, the average and the standard deviation of Z are calculated as the HR and SDNN.

Second, we apply voting algorithm to generate the rule-modulated map $R = \{R_t\}$, where $R_t$ here is defined as the possibility of R-peaks at time point $t$ according to votes from supporting regions. The supporting region is defined according to the knowledge about heart rhythm. As we know, R-peak occurrences, which are approximately-integral multiples of HR away from the current time point, can be used as supporting evidences of the possibility of the R-peak occurrence at the current time point. Therefore, the supporting regions consist of the time points which are approximately-integral multiples of HR away. Here we denote the supporting regions as $\{\phi(c_i, l_i)\}$ where $c_i$ and $l_i$ are the centers and widths of each region. The centers are set to $\{t \pm k \cdot HR\}$, where $k$ represents the $k$-th center before or after the time point $t$. $l_i$ can be estimated according to SDNN and the three-sigma rule of thumb as:

$$l_i = 6 \cdot \sqrt{k} \cdot SDNN \quad (5)$$

With a constraint that the supporting regions should not be overlapped with each other, we can determine how many supporting regions could be used for voting. For simplicity, we set the maximum of the number of supporting regions as six.

$R_t$ is then calculated as:

$$R_t = \frac{\sum_i \left( w_i \cdot \max\left(\phi(c_i, l_i)\right) \big/ l_i \right)}{\sum_i \left( 1 / l_i \right)} \quad (6)$$

where $w_i$ denotes the confidence about the vote from each region:

$$w_i = \frac{1}{\sqrt{2\pi} \cdot l_i} \cdot e^{(\arg\max(\phi(c_i, l_i)) - c_i) / 2 \cdot l_i^2} \quad (7)$$

If no supporting regions can be found due to arrhythmia or insufficient time intervals of R-peaks, we set $R_t = 0$.

## EXPERIMENTAL SETTING

This section sets up experiments to evaluate ReNN in the task of R-peak detection from ECG. ECG has been used as the firsthand information to analyze the structure and function of the heart for about 100 years [27], and it is becoming popular in smart wearable devices to monitor the status of users [28, 29]. However, it is still challenging to accurately detect R-peaks due to several types of noises affecting their occurrences, such as motion artifact, muscular activation and AC interference [29].

### A. Dataset

In the experiments, we use an ECG dataset constructed by LOHAS Tech., where the ECG data are measured at sampling rate of 125 Hz with a single-lead device between electrodes placed on left and right arms. The dataset includes 16600 ECG measurements from 684 users as well as the labels of R-peaks in each ECG measurement. Most of the users are healthy with normal sinus rhythm. Each ECG measurement is about 12 seconds long, which is filtered with a high-pass filter at 1Hz and a low-pass filter at 32Hz to remove irrelative signal components.

The dataset includes about 260k labels of R-peaks. The labels are first labeled by the PT algorithm [30], and then suspicious labels are carefully checked and re-labeled by human experts. The standard deviation of R-peak intervals (referred as SDNN) is used to estimate the heart rate variability for each ECG measurement. 16036 (97%) ECG measurements in the dataset are with low SDNN (<0.06s), and only 26 ECG measurements are with high SDNN (>0.1s).

We split the dataset into a training dataset and a testing dataset which include 80% and 20% of the ECG measurements, respectively. They are used for model training and evaluation.

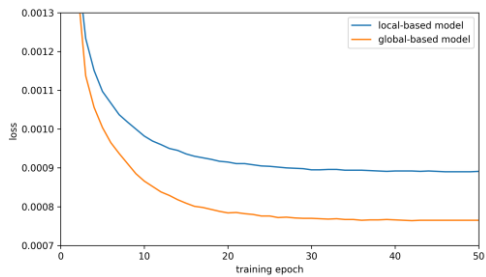

Figure 3. Loss curves in the training process of the local based model and the ReNN model. Horizontal axis represents the number of training epochs, and the vertical axis represents the value of the loss function.

### B. Model training

To train ReNN model with FCN block, we use the Adam optimizer with the weighted cross-entropy between the labels and the model output as loss function. We feed one ECG measurement and its labels of R-peaks and apply the optimizer at each training step. All the ECG measurements are fed one by one for 100 epochs, and thus each measurement is fed for 100 times. The learning rate is set as 0.0001 at the beginning, and then decayed every 1000 steps exponentially at a rate of 0.99. The total number of training step is about 1.3 million.

### C. Model Evaluation

ReNN is compared with the basic FCN method (referred as local-based inference). The number of channels is adjusted to make a balance computational complexity and performances. We use the value of loss function to analyze the convergence of model training process. To evaluate the detection performance, we use the following metrics: the number of R-peaks correctly detected (true-positives, TP), the number of R-peaks missed (false negatives, FN), and the number of noise samples wrongly detected as R-peaks (false-positives, FP). We also calculate the F1-score ( $2 \cdot TP / (2 \cdot TP + FP + FN)$ ) which is the harmonic mean of precision and sensitivity of a detection model.

Given that the exact location of a R-peak may be unstable between two neighboring sampling-points due to noises, we set a tolerance interval for the detection. That is, a correctly detected R-peak is judged if and only if the model outputs a positive during the tolerance interval of a labeled R-peak. The tolerance interval is set as the length of two sampling points, i.e. 16ms.

## RESULTS

### A. Convergence

Fig. 3 shows the convergence of loss function at the early training phase. The loss curves of the local-based model and the ReNN model decrease rapidly during the first 20 training epochs, and then tend to be stable gradually. The final values of the loss function are 8.904e-4 and 7.648e-4 for the local-based model and ReNN model, respectively. The results reveal that both loss functions have converged on the training dataset, and the ReNN model can achieve better fitting performance.

### B. Detection performance

Table 1 lists the detection performances of the local-based inference and the global-based inference (ReNN) on the test dataset.

Table 1. Detection performances of the Local-based inferences and Global-based inferences on the test dataset. #C represents the number of channels in the FCN model.

| Type | #C | TP | FP | FN | F1-score |
|---|---|---|---|---|---|
| Local | 4 | 50212 | 249 | 151 | 0.9960 |
| | 8 | 50299 | 116 | 64 | 0.9982 |
| | 16 | 50308 | 83 | 55 | 0.9986 |
| | 32 | 50317 | 65 | 46 | 0.9989 |
| | **64** | **50323** | **33** | **40** | **0.9993** |
| Global | 4 | 50294 | 73 | 69 | 0.9986 |
| | **8** | **50346** | **31** | **17** | **0.9995** |
| | 16 | 50330 | 33 | 33 | 0.9993 |
| | 32 | 50319 | 37 | 44 | 0.9992 |
| | 64 | 50321 | 25 | 42 | 0.9993 |

According to the F1-score, ReNN achieves low error rate when the number of channels is only eight, while the local-based inference achieves comparable performance with 64 channels. However, the computational cost of local-based inference with 64 channels is about several times of that of ReNN. The results reveal that ReNN can make use of the rule-modulated map to enhance the accuracy to detect the R-peaks. The results also reveal that ReNN has the potential to reduce computational cost of deep neural networks.

Fig. 4 illustrates an example of ECG measurements and ReNN detection involving the output of feature-mapping block, rule-modulating block, and global-mapping block. With the voltage range of only 0.2 mV, the signal of this ECG measurement is weaker than standard ECG where the voltage range is about 1 mV. There are also some noises that might be wrongly recognized as R-peaks.

Intuitively, it is hard for us to distinguish the R-peaks from the signal with only local patterns. Nevertheless, we have knowledge about the heart rhythm of the user. This ECG is measured from a 25-year-old female who is healthy and with no history of heart diseases. Accordingly, there should be no premature beat or other abnormal beat in this ECG measurement, and thus we can discriminate the R-peaks from the noisy signal with heart rate and neighboring R-peaks.

The above analyzing process for human from local patterns to global patterns, i.e. HR has also been reflected in the detection process of ReNN. Firstly, the feature-mapping block detects potential R-peaks according to local patterns in the receptive fields. Secondly, the rule-modulating block analyzes heart rhythm from the feature maps, and then estimates distributions of heartbeats based on the heart rhythm and features at neighboring heartbeats. Finally, the global-mapping block synthesizes the original ECG and the rule-modulated map to refine the detection. In the example shown in Fig. 4, we can find that a noisy peak (false positive by local-based model) has been suppressed successfully, and a distorted R-peak has been enhanced in the global-based inference.

### C. Interpretability

To interpret an inference, we should ask and answer the following questions:

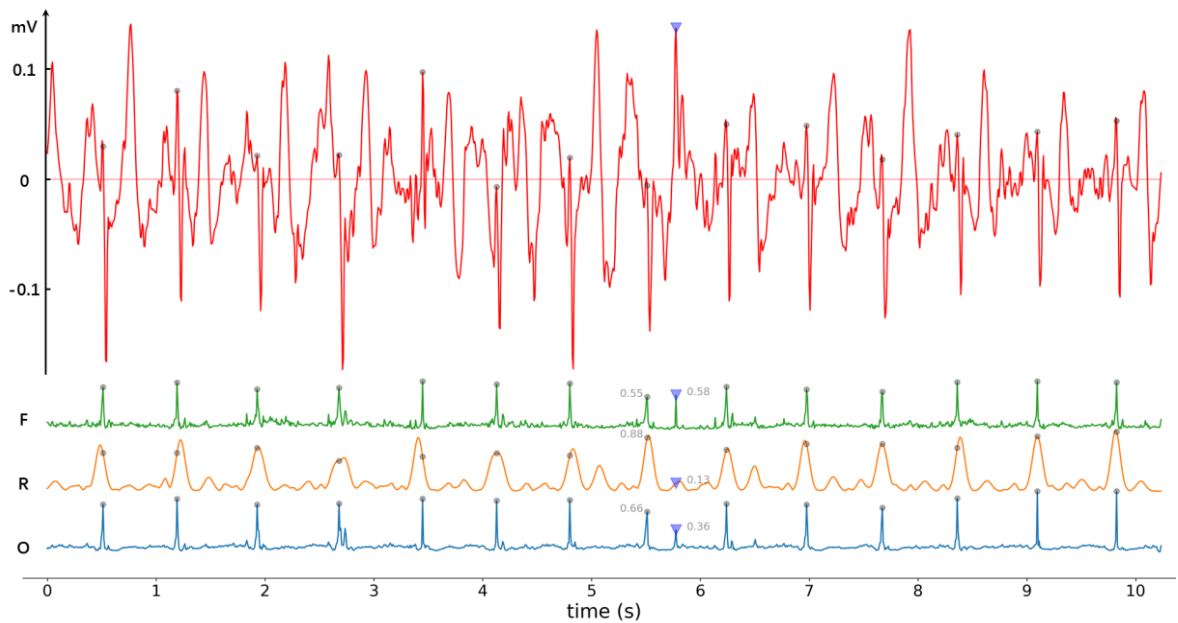

Figure 4. ReNN detection on an ECG measurement. The top line is the input time-series data of the ECG measurement, where the ECG signals are very weak. The other three lines from top to bottom are the outputs of feature-mapping block (line-F), rule-modulating block (line-R) and global-mapping block (line-O), respectively. The four lines are aligned according to time axis. The solid circles anchored on the lines are the time points labeled as R-peaks. The downward triangle shows a false positive (FP) by local-based inference (high value on line-F), while global-based inference (lower value on line-O) reduces the probability of R-peak at this time point due to little support from heart rhythm (low value on line-R). Besides, the first R-peak in front of the triangle is distorted due to noise. It is detected with higher probability with the support from heart rhythm (high value on line-R).

➢ What are the local patterns, and what are rules among them?

➢ What are the local-based inferences with only local patterns?

➢ What are the global-based inferences with the help of rules?

➢ How can we explain the conflicts between the local-based and the global-based inferences?

We can answer these questions with the example in Fig. 4:

✓ The local patterns are local morphologies of the R-peaks, which are detected by the feature-mapping block. The rules among the R-peaks are designed according to HR and SDNN. Furthermore, we can disassemble the local morphologies into more details, i.e. the P-Q-R-S-T peaks, and train feature-mapping block for the details. Then, rules about the order and time constraints about these peaks can be applied.

✓ Line-F shows the local-based inferences that only use local patterns. Some suspicious R-peaks are detected according to the local-based inference.

✓ Line-O shows the global-based inferences which combine local patterns and rules. Line-O has less jitters than Line-F, indicating that the combination of local patterns and rules have reduced the uncertainty of global-based inferences.

✓ We should check and analyze the conflicts between the local-based and the global-based inferences carefully. There might be two possibilities. Firstly, the conflicts may indicate exceptions of the rules, such as a premature beat of the heart, which have the value of diagnosis if it is a new case. If some indications for the conflicts have already been diagnosed, we should add a label to the user and suppress similar conflicts, or we should take some actions such as medication. Secondly, noise may cause the conflicts, which indicate that current ECG measuring and analyzing system is not efficient to discover real abnormalities of such kind. The usage should be checked, or the system should be improved.

The above question-answering is similar to the inference procedure of human, leading to better interpretability for ReNN.

## DISCUSSION

This paper has proposed ReNN to overcome the limitation of current ANNs – lack of knowledge, and further validated this approach in an ECG R-peak detection problem. By introducing knowledge with the rule-embedding approach, ReNN could improve detection accuracy as well as reduce model complexity. The needs of big dataset for training can be mitigated since the model complexity is reduced. Besides, ReNN has improved the interpretability of the ANN-based technologies. The global-based inference of ReNN can be interpreted as the interaction between local patterns and rules among the local patterns.

Modeling of long-term dependencies is a difficult problem in machine learning area, as the uncertainty grows rapidly when the input data become higher-dimensional. The current popular solution is to train very deep ANN with very big dataset, such as LSTM [31] and ResNet [1]. The attention mechanism [32, 33] also plays a role in modeling long-term dependencies into a context vector, and the price is also big data. We argue that the rule-embedding approach could provide a new solution to make use of long-term dependencies efficiently, especially when we have already accumulated some knowledge about the long-term dependencies. It might become a new way for human-computer interaction where local-pattern-rules are the contents of interaction. Computers can tell us what local patterns they have learnt and observed, and we can teach them what rules are there among the local patterns. It seems just like that we are teaching a kid or a student, and thus the computers seem more intelligent.

There might be a question that why the rules should be embedded in ANN rather than be placed ahead or behind. Rules are not placed ahead because most of our knowledge is about some semantic features extracted from the raw data. The hidden layers of ANN can play the role to extract the features. Rules are not placed behind because we want to model the dependencies between local patterns and the rules automatically and to make inferences synthetically. Rules just play a role to extract global features, i.e. the rule-modulated map, and thus we can add rules freely without worry about conflicts between rules or combinatorial explosion of multiple rules.

Besides, Rules in ReNN are independent of the optimization procedure, so human experts with domain knowledge could design rules freely without constraints of differentiability. This makes the combination of ANN and existing knowledge much easier. Furthermore, rules could be accumulated and organized with cross-validation or the probabilistic graphical model based reasoning systems [34, 35], leading to a rule base where rules are marked with applicable tasks or domains. With the rule base, computers could gradually know how and when to apply what rules. This would be helpful in transfer learning where very few samples can be used or unsupervised learning where no labels can be used. Such kind of human-computer interaction might lead a new thinking about artificial intelligence.

## ACKNOWLEDGMENT

The author thanks Prof. Jue Wang from Institute of Automation, Chinese Academy of Sciences for introducing the philosophy of "structure + average" for AI, which has inspired the ReNN. The author thanks Dr. Jidong Liu from Shandong Provincial Hospital for help to build the ECG dataset, and Yao Wang and Yudong Zhu from LOHAS Tech. for their supports, and Zongbo Zhang, Shuyu Han, Dongyang Mei, Ran Yan, Yuzhi Mu and Shuai Ma for weekly discussion of AI.